\documentclass{article}
\usepackage{spconf,amsmath,amssymb,graphicx}
\usepackage{booktabs,multirow}
\usepackage{xcolor}
\usepackage{url}
\usepackage{pifont}
\usepackage[protrusion=false]{microtype}
\usepackage[hidelinks,hyperfootnotes=false]{hyperref} 
\newcommand{\envelope}{\mbox{\footnotesize\ding{41}}}

\newcommand{\ours}{SurgStream}
\newcommand{\bench}{SurgStream-Bench}
\newcommand{\tok}[1]{\textit{#1}}
\newcommand{\cmark}{$\checkmark$}

\def\L{L}
\makeatletter
\renewcommand\section{\@startsection{section}{1}{\z@}{-2.0ex \@plus -0.5ex \@minus -0.2ex}{1.0ex \@plus 0.2ex}{\normalfont}}
\renewcommand\subsection{\@startsection{subsection}{2}{\z@}{-1.6ex \@plus -0.4ex \@minus -0.2ex}{0.7ex \@plus 0.2ex}{\normalfont}}
\makeatother

\newcommand{\titlelogo}{\includegraphics[height=0.5in]{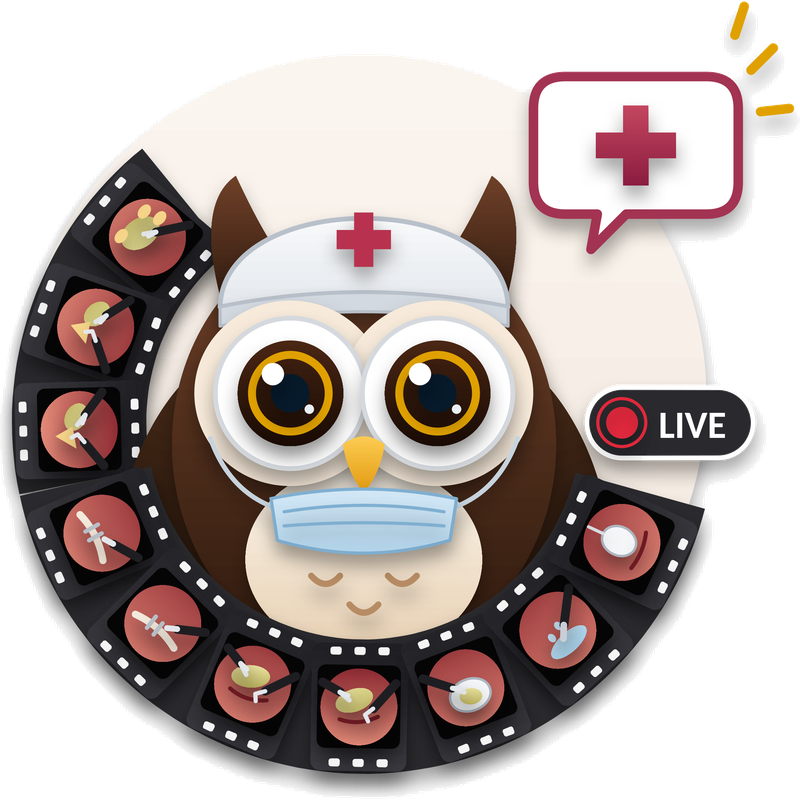}}
\newcommand{\titlename}{\textcolor[RGB]{196,40,78}{S}\textcolor[RGB]{187,39,77}{U}\textcolor[RGB]{179,37,75}{R}\textcolor[RGB]{170,36,74}{G}\textcolor[RGB]{161,35,73}{S}\textcolor[RGB]{153,33,71}{T}\textcolor[RGB]{144,32,70}{R}\textcolor[RGB]{135,31,69}{E}\textcolor[RGB]{127,29,67}{A}\textcolor[RGB]{118,28,66}{M}}
\newcommand{\titlewithlogo}[2]{\raisebox{-0.5\height}{\titlelogo}\hspace{0.09in}\raisebox{-0.5\height}{\fontsize{11.3}{14}\selectfont\begin{tabular}{@{}c@{}}#1\\ #2\end{tabular}}}
\makeatletter\def\@maketitle{\newpage\null\vskip 0.2em\begin{center}{\large\bf \@title\par}\vskip 1.1em{\large\lineskip .5em\begin{tabular}[t]{c}\@name \\ \@address\end{tabular}\par}\end{center}\par\vskip 1.0em}\makeatother
\title{\titlewithlogo{\titlename: A Workflow-Anchored Proactive Streaming Framework}{and Benchmark for Real-Time Surgical Video Understanding}}

\name{Ruipeng Ma$^{1,2,*}$, Qing Xu$^{3,*}$, Jie Liang$^{2,\envelope}$, Zhen Chen$^{1,\envelope}$}
\address{$^{1}$Department of Data Science and Artificial Intelligence, The Hong Kong Polytechnic University\\
    $^{2}$Eastern Institute of Technology, Ningbo\quad $^{3}$University of Nottingham, UK}

\begin{document}
\microtypesetup{expansion=false}
\ninept
\sloppy
\maketitle

\begin{abstract}
Surgical video understanding is essential for intraoperative assistance such as workflow monitoring, instrument preparation and procedural documentation. Delivering such assistance in real time requires a model to jointly perceive the ongoing scene, remember past surgical events, and respond before clinically relevant moments pass. Existing approaches provide limited support for jointly retaining procedural evidence and deciding when to respond in streaming surgical video. In this work, we propose \ours{}, a workflow-anchored proactive streaming framework for real-time surgical video understanding. We first introduce Workflow-Anchored Memory (WAM) to record structured instrument--action--target observations and fold them into phase-level summaries, providing compact and queryable surgical memory. Deadline-aware Transition Learning (DTL) is then devised to supervise response states within clinical windows, enabling timely responses before procedural deadlines. We further introduce \bench{}, comprising 11.7K streaming interactions across various tasks covering real-time perception, backward memory queries, forward proactive QA, proactive alerts and caption memory. Experiments on \bench{} show gains over state-of-the-art methods, especially in memory and forward proactive QA. The source code will be available.
\end{abstract}


{\let\thefootnote\relax\footnotetext{$^{*}$ Equal contribution.\quad $^{\envelope}$ Corresponding authors.}}

\section{Introduction}
\label{sec:intro}

\begin{figure*}[t]
  \centering
  \includegraphics[width=\linewidth]{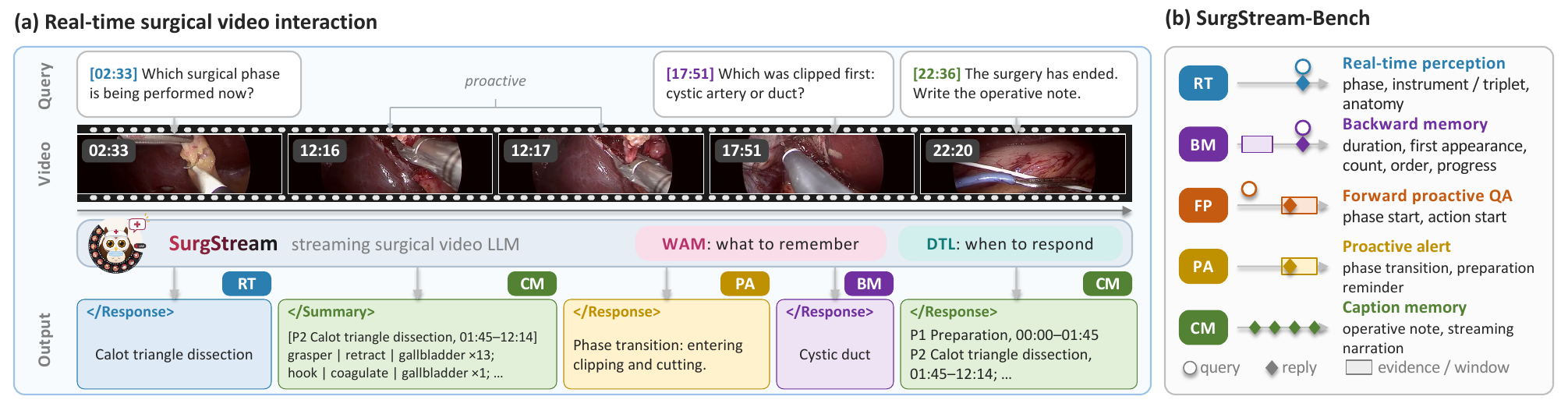}
  \caption{Illustration of \ours{} and \bench{}. (a) An example of streaming surgical interaction, including current-scene questions, retrospective queries, proactive alerts and memory writing. (b) The five temporal modes evaluated by \bench{}.}
  \label{fig:teaser}
\end{figure*}

Real-time surgical video understanding plays a crucial role in intraoperative assistance~\cite{maierhein2017sds,maierhein2022sds}, such as phase-aware instrument preparation, procedural documentation and progress tracking. Such assistance involves understanding the current scene, recalling past events and delivering responses within relevant procedural windows. Fig.~\ref{fig:teaser} illustrates these interaction requirements and the temporal modes considered in this work. Surgical videos are long, visually homogeneous and densely structured by the workflow~\cite{twinanda2017endonet}. For time-sensitive assistance, delayed responses may lose their practical value, making real-time surgical interaction challenging.

Early surgical video-language models primarily focused on visual question answering on images or complete videos~\cite{seenivasan2022surgicalvqa,yuan2024ssgvqa,he2024pitvqa,li2024llavasurg}. Subsequently, general and medical vision-language models~\cite{bai2025qwen3vl,wang2025internvl35,jiang2025hulumed} achieved significant progress in visual understanding, and real-time surgical commentary~\cite{he2026surgonair} narrated the current frames on the fly. More recently, streaming video large language models (LLMs)~\cite{chen2024videollmonline,wang2024mmduet,qian2025dispider,chen2025livecc,xu2025streamingvlm} have enabled causal perception and proactive response on continuous video streams, and OneStreamer~\cite{zeng2026onestreamer} further unifies perception, memory and proactive response in a single model.

Despite this progress, existing methods are not explicitly designed around surgical workflows. Free-text memory does not explicitly preserve phase boundaries or structured event counts, making retrospective queries difficult to support reliably. Response supervision at a single annotated time~\cite{zeng2026onestreamer} also leaves the interval of acceptable responses implicit. These limitations motivate workflow-structured memory, supervision within valid response windows, and evaluation of both answer content and timing.

To address these limitations, we propose \ours{}, a workflow-anchored proactive streaming framework that integrates compact surgical memory with response supervision within valid procedural windows. For memory construction, we introduce Workflow-Anchored Memory (WAM) to record stable instrument--action--target changes and consolidate each completed phase into a compact \textit{Summary}, preserving evidence for retrospective queries. We further devise Deadline-aware Transition Learning (DTL) to resample response targets within label-derived clinical windows in each epoch, encouraging early responses once evidence is available and before procedural deadlines. Moreover, we construct \bench{}, spanning diverse tasks across five clinical scenarios and five temporal modes. Extensive experiments demonstrate that \ours{} outperforms competing methods, with particularly strong gains in memory and forward proactive QA.

\section{Methodology}
\label{sec:method}

\begin{figure*}[t]
  \centering
  \includegraphics[width=\linewidth]{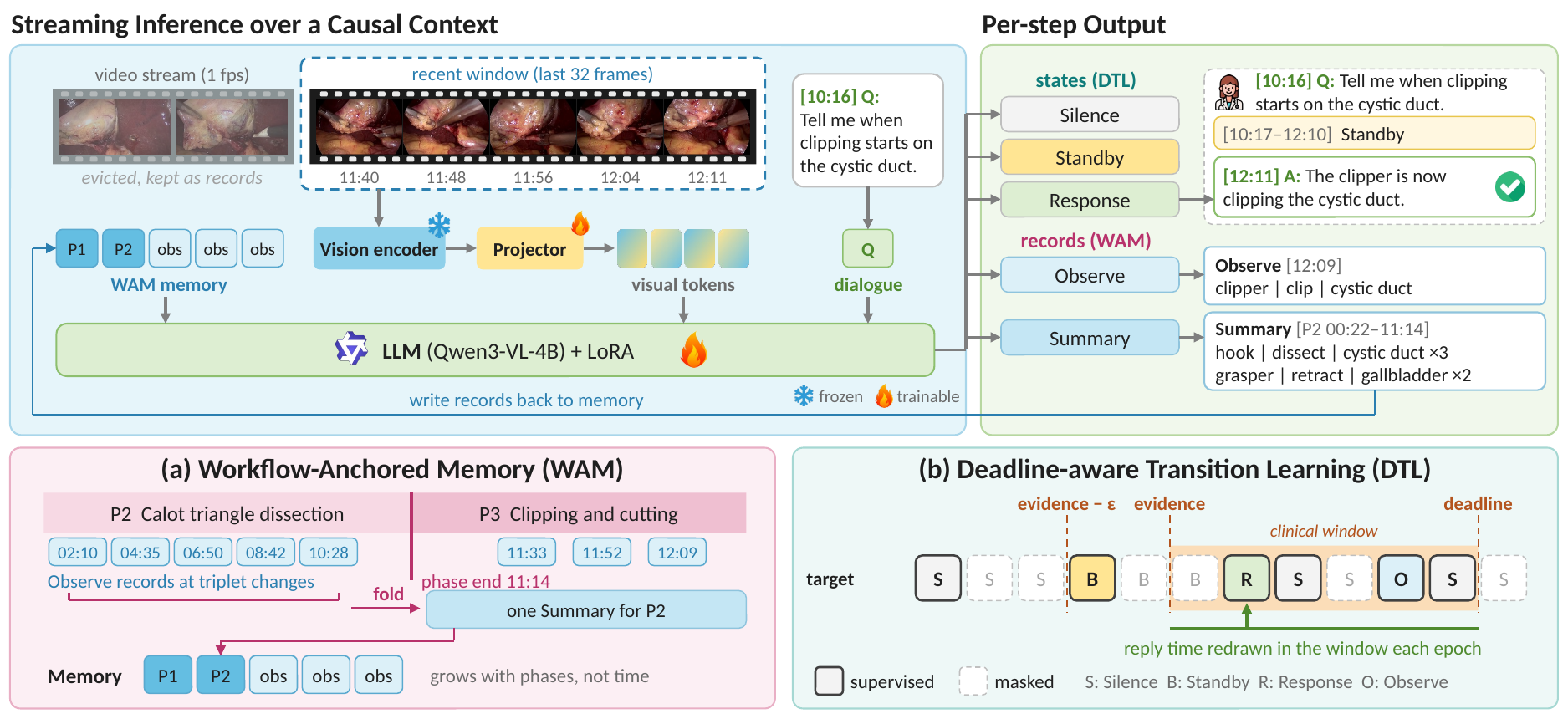}
  \caption{Overview of \ours{}. At each step, the LLM reads WAM records, the recent visual window and pending dialogue, and emits a control state with optional text. (a) WAM writes \tok{Observe} records at triplet changes and folds them into a \tok{Summary} at each phase boundary. (b) DTL redraws the response time within the clinical window in each epoch and supervises states on a sparse~mask.}
  \label{fig:overview}
\end{figure*}

As shown in Fig.~\ref{fig:overview}, \ours{} performs real-time surgical video interaction within a single streaming video LLM. Given a video stream $\{x_t\}_{t=1}^{T}$ arriving at 1\,fps, the model reads at each step $t$ the causal context $C_t=[M_t;V_t;D_t]$, consisting of the memory $M_t$, the recent visual window $V_t=\{x_{t-W+1},\dots,x_t\}$ and the dialogue $D_t$, and emits a control state $s_t\in\{$\tok{Silence}, \tok{Standby}, \tok{Response}, \tok{Observe}, \tok{Summary}$\}$ followed by optional text. WAM and DTL anchor memory construction and response timing to the surgical workflow.

\subsection{Workflow-Anchored Memory}
\label{ssec:wam}
To provide compact and queryable surgical memory, we propose WAM, which anchors both the write points and the content of memory to the surgical workflow instead of letting the model decide the boundaries of free-text captions. For training-target construction, let $u_t$ denote the structured surgical state at time $t$, \emph{i.e.}, the set of instrument--action--target triplets or present instruments. When $u_t$ changes and the new state remains stable for $\kappa$ seconds, the model writes an \tok{Observe} record $o_\tau=\langle t,\Delta u_t\rangle$ at $\tau=t+\kappa-1$, where $\Delta u_t$ denotes the elements newly appearing at time $t$ relative to the preceding state. The record lists these elements as ``[mm:ss] instrument \textbar{} verb \textbar{} target''. Let $O_k$ collect the \tok{Observe} records $o_\tau$ whose event times $t$ fall within phase $k$, $[b_k,b_{k+1})$. After the phase boundary $b_{k+1}$ is confirmed, the model writes a \tok{Summary} record $\sigma_k=\langle p_k,[b_k,b_{k+1}),\{(c,n_k(c))\}\rangle$, where $p_k$ is the phase label and $n_k(c)$ counts the onsets of element $c$ in $O_k$ (\emph{e.g.}, clip$\times$3). We replace $O_k$ with $\sigma_k$ and retain the other memory records $R_t$:
\begin{equation}
\begin{alignedat}{2} & R_t &&= \{\,m\in M_t \mid m\notin O_k\,\},\\ & M_{t+1} &&= R_t\cup\{\sigma_k\}. \end{alignedat}
\label{eq:fold}
\end{equation}
WAM retains phase summaries for completed phases and detailed observations for the current phase, reducing the accumulation of historical records. Phase intervals and event counts support retrospective queries, while retained \tok{Observe} records preserve precise event times. Expert annotations define these training targets. At inference, the model generates memory records solely from causal video context. The memory loss $\L_{\mathrm{mem}}$ is token-averaged cross-entropy conditioned on the state and causal context.

\begin{table*}[!t]
\vspace{-8pt}
\centering
\caption{Comparison on \bench{} (\%). $^\dagger$: fine-tuned. Proactive: the model autonomously decides when to respond. Periodic: the model is queried every 5\,s. \textbf{Bold} marks the better result in each pair.}
\label{tab:main}
\scriptsize
\setlength{\tabcolsep}{5pt}
\renewcommand{\arraystretch}{1.15}
\setlength{\aboverulesep}{0.25ex}
\setlength{\belowrulesep}{0.25ex}
\resizebox{\textwidth}{!}{\begin{tabular}{lcc|cccccc|cccccc}
\toprule
\multirow{2}{*}{Methods} & \multirow{2}{*}{Size} & \multirow{2}{*}{Response mode} & RT & BM & FP & PA & CM & \multirow{2}{*}{Avg$_\mathrm{M}\!\uparrow$} & \multirow{2}{*}{S1} & \multirow{2}{*}{S2} & \multirow{2}{*}{S3} & \multirow{2}{*}{S4} & \multirow{2}{*}{S5} & \multirow{2}{*}{Avg$_\mathrm{S}\!\uparrow$} \\
 & & & Acc$\uparrow$ & Score$\uparrow$ & T-Acc$\uparrow$ & T-F1$\uparrow$ & S-F1$\uparrow$ & & & & & & & \\
\midrule
Qwen3-VL~\cite{bai2025qwen3vl} & 4B & \multirow{8}{*}{Periodic} & 41.73 & 29.93 & 0.43 & 0.30 & 4.26 & 15.33 & 12.11 & 30.13 & 46.95 & 0.68 & 8.52 & 19.68 \\
Qwen3.5~\cite{qwen2026qwen35} & 4B & & 42.70 & 27.77 & 2.69 & 1.08 & 5.98 & 16.04 & 13.06 & 28.88 & 43.46 & 5.38 & 7.93 & 19.74 \\
Gemma~4~\cite{gemma2026gemma4} & E4B & & 42.89 & 19.88 & 0.86 & 1.60 & 6.62 & 14.37 & 14.17 & 13.06 & 52.39 & 4.64 & 9.08 & 18.67 \\
MedGemma~\cite{sellergren2026medgemma15} & 4B & & 34.54 & 11.14 & 0.43 & 0.73 & 3.48 & 10.06 & 8.84 & 11.04 & 36.66 & 0.89 & 5.17 & 12.52 \\
SurgVidLM~\cite{wang2025surgvidlm} & 7B & & 36.90 & 17.93 & 0.00 & 0.00 & 0.62 & 11.09 & 12.23 & 10.22 & 49.16 & 0.00 & 1.24 & 14.57 \\
Hulu-Med~\cite{jiang2025hulumed} & 4B & & 42.77 & 29.70 & 0.00 & 0.78 & 5.95 & 15.84 & 13.18 & 30.15 & 42.23 & 6.01 & 7.28 & 19.77 \\
MedVLM~\cite{su2026medgrpo} & 7B & & 55.35 & 27.80 & 0.54 & 1.84 & 2.43 & 17.59 & 13.75 & 33.36 & 53.13 & 0.68 & 4.85 & 21.16 \\
EndoChat~\cite{wang2025endochat} & 13B & & 39.62 & 27.92 & 0.00 & 0.21 & 0.86 & 13.72 & 5.29 & 29.57 & 55.37 & 2.33 & 0.48 & 18.61 \\
\midrule
Qwen3-VL\smash{$^\dagger$}~\cite{bai2025qwen3vl} & 4B & \multirow{2}{*}{Periodic} & 82.04 & 52.33 & 21.84 & \textbf{24.37} & 20.68 & 40.25 & 35.04 & 62.50 & 71.87 & 22.95 & \textbf{26.73} & 43.82 \\
\textbf{\ours{}} & 4B & & \textbf{86.26} & \textbf{58.62} & \textbf{27.43} & 20.71 & \textbf{22.83} & \textbf{43.17} & \textbf{35.44} & \textbf{71.19} & \textbf{77.93} & \textbf{25.15} & 25.68 & \textbf{47.08} \\
\midrule
OneStreamer\smash{$^\dagger$}~\cite{zeng2026onestreamer} & 4B & \multirow{2}{*}{Proactive} & 83.27 & 59.10 & 10.87 & 14.78 & 15.16 & 36.64 & 32.99 & 62.72 & 74.81 & 18.49 & 24.47 & 42.70 \\
\textbf{\ours{}} & 4B & & \textbf{84.07} & \textbf{64.13} & \textbf{23.13} & \textbf{16.64} & \textbf{25.22} & \textbf{42.64} & \textbf{36.84} & \textbf{67.70} & \textbf{78.06} & \textbf{31.45} & \textbf{28.79} & \textbf{48.57} \\
\bottomrule
\end{tabular}}
\vspace{-8pt}
\end{table*}

\subsection{Deadline-aware Transition Learning}
\label{ssec:dtl}
To enable timely responses within valid clinical windows, we propose DTL, which extends generic sparse supervision~\cite{zeng2026onestreamer} beyond a single annotated response time. Each response target $j$ is assigned a clinical window $[a_j,d_j]$, where $a_j$ is the earliest admissible response time and $d_j$ is the deadline, both derived from annotations and task-specific timing rules (\emph{e.g.}, a preparation reminder is due at the next phase onset). To favour early responses, we resample the target time within the valid window at each epoch using inverse transform sampling. Let $U_j\sim\operatorname{Uniform}(0,1)$ and $\Delta_j=d_j-a_j+1$. \mbox{The sampled delay $\tau_j$ and response time $r_j$ are:}
\begin{align}
\tau_j&=-\eta\log\!\left[1-U_j\left(1-e^{-\Delta_j/\eta}\right)\right],\label{eq:sample}\\
r_j&=\min\!\left\{d_j,\ a_j+\lfloor\tau_j\rfloor\right\}.\label{eq:density}
\end{align}
Here, timestamps are discretized at one-second intervals, and $\eta>0$ controls the decay of the sampling distribution. We assign $s^*_{r_j}=\text{\tok{Response}}$, $s^*_t=\text{\tok{Standby}}$ for $t\in[a_j-\epsilon,r_j)$, and \tok{Silence} elsewhere. Following~\cite{zeng2026onestreamer}, the supervised set $\Omega$ retains output anchors (including the resampled $r_j$) and uses quota-based sampling over state-transition groups to select other positions. The state loss is:
\begin{equation}
\L_{\mathrm{DTL}}=-\frac{1}{|\Omega|}\sum_{t\in\Omega}\log\pi_\theta(s^*_t\mid C_t).
\label{eq:ldtl}
\end{equation}
Keeping $\Omega$ sparse limits the contribution of abundant non-response positions to the state loss. Redrawing $r_j$ in each epoch with earlier times favoured by Eq.~(\ref{eq:sample}) encourages the model to respond soon after the earliest admissible time while respecting the clinical deadline.

\ours{} is trained jointly with the cross-entropy of the reply text, $\L_{\mathrm{mem}}$ and $\L_{\mathrm{DTL}}$, weighted equally. At inference, the states are emitted in the order \tok{Summary}$\rightarrow$\tok{Observe}$\rightarrow$\tok{Response} until \tok{Silence} or \tok{Standby}. Inference uses a fixed visual window and retained memory.

\subsection{\bench{}}
\label{ssec:bench}
We construct \bench{} by converting expert annotations from Cholec80~\cite{twinanda2017endonet} and CholecT50~\cite{nwoye2022rendezvous} into streaming interactions. Phase, instrument and action-triplet labels are aligned along each procedure. Shared recordings are identified using the official overlap mapping\footnote{\url{github.com/CAMMA-public/camma_dataset_overlaps}}, represented once with merged labels, and assigned to procedure-level splits to prevent leakage. \par We organize the interactions into five temporal modes: real-time perception (RT), backward memory (BM), forward proactive QA (FP), proactive alert (PA) and caption memory (CM). Across these modes, tasks cover five clinical scenarios: workflow (S1), instrument and action (S2), anatomy (S3), documentation (S4) and narration (S5). Current-scene questions use labels at the query time. Past labels support queries on durations, first appearances, counts and order. Phase and triplet labels yield summaries and operative notes. \par Proactive targets pair an expected answer with an earliest admissible response time and a deadline. Phase-transition alerts follow annotated transitions, while preparation reminders precede the next phase. This construction evaluates both whether the answer is correct and whether it arrives within the task-specific response window. \par RT and BM use accuracy or task-specific tolerances. FP timing accuracy (T-Acc) checks the first reply after the query for correct content within $[a_j,d_j]$. PA timing F1 (T-F1) uses one-to-one temporal matching with the same requirements. CM uses structured F1 (S-F1) over phase intervals and triplet events. Exact vocabulary matching avoids an LLM judge. Avg$_\mathrm{M}$ and Avg$_\mathrm{S}$ average mode and scenario scores, each computed as the mean of its task scores.

\begin{table}[t]
\centering
\caption{Ablation study under proactive inference.}
\label{tab:ablation}
\footnotesize
\setlength{\tabcolsep}{4pt}
\resizebox{\columnwidth}{!}{\begin{tabular}{cc|ccccccc}
\toprule
WAM & DTL & RT$\uparrow$ & BM$\uparrow$ & FP$\uparrow$ & PA$\uparrow$ & CM$\uparrow$ & Avg$_\mathrm{M}\!\uparrow$ & Avg$_\mathrm{S}\!\uparrow$ \\
\midrule
 & & 83.27 & 59.10 & 10.87 & 14.78 & 15.16 & 36.64 & 42.70 \\
\cmark & & 82.74 & 62.46 & 14.53 & 17.58 & 22.84 & 40.03 & 46.74 \\
 & \cmark & 83.48 & 61.97 & 20.23 & 14.37 & 22.34 & 40.48 & 45.96 \\
\cmark & \cmark & 84.07 & 64.13 & 23.13 & 16.64 & 25.22 & 42.64 & 48.57 \\
\bottomrule
\end{tabular}}
\end{table}

\begin{figure*}[t]
  \centering
  \includegraphics[width=\linewidth]{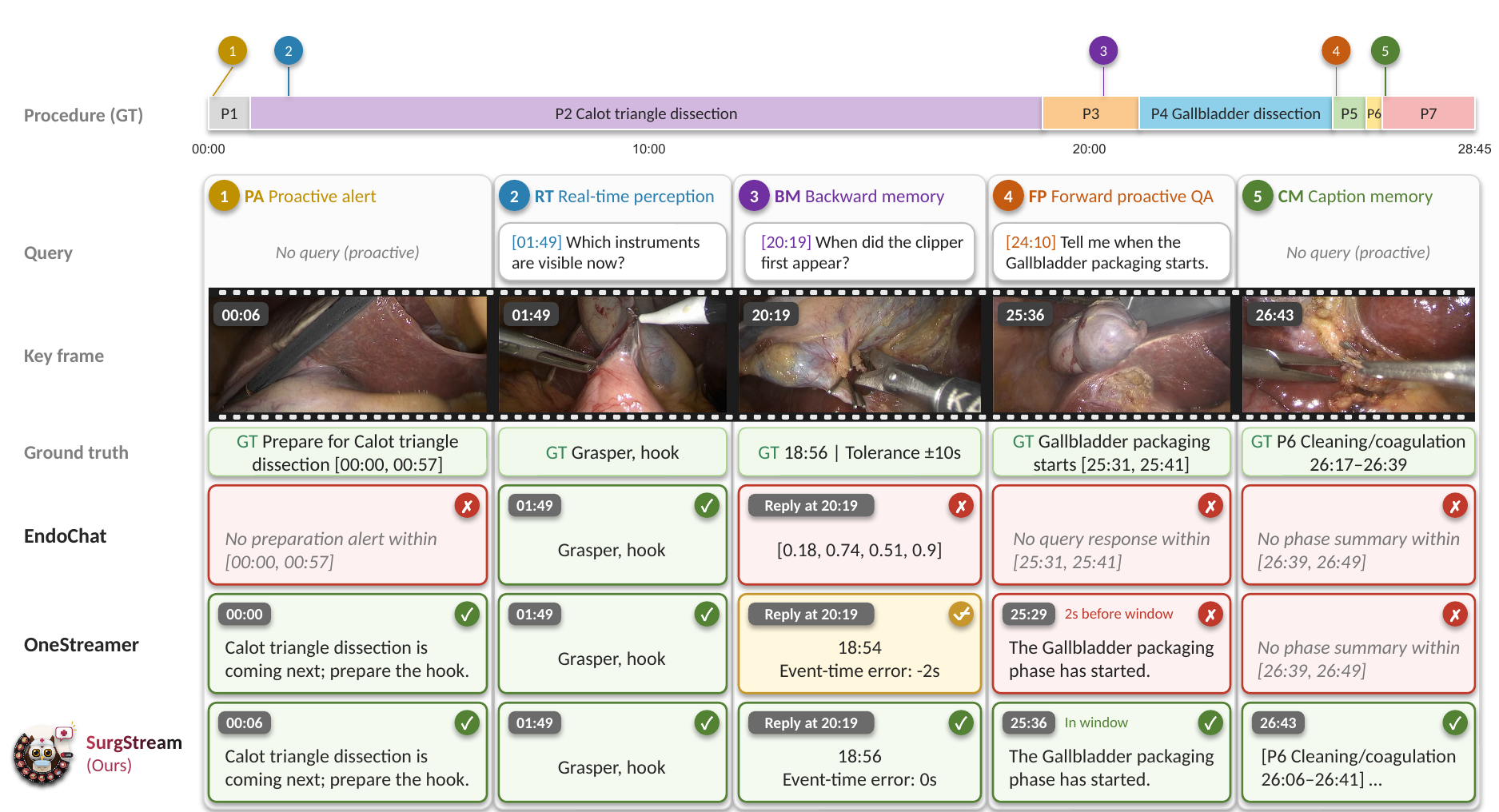}
  \caption{Qualitative comparison of EndoChat, fine-tuned OneStreamer and \ours{} on one test procedure across five temporal modes. Responses are excerpted. BM compares event-time estimates ($\pm10$\,s tolerance), and the CM checkmark denotes a matched phase record.}
  \label{fig:qualitative}
\end{figure*}

\begin{figure}[t]
\centering
\includegraphics[width=\columnwidth]{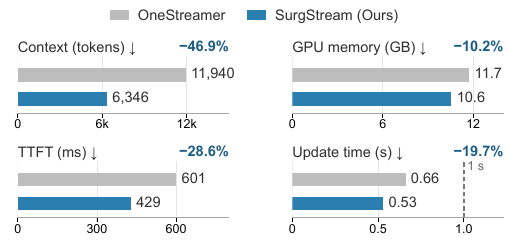}
\caption{Efficiency comparison of \ours{} and OneStreamer on two full-length test procedures at 1\,fps. Update time is averaged over video steps. GPU memory is reported in GB.}
\label{fig:eff}
\end{figure}

\section{Experiments}
\label{sec:exp}

\subsection{Dataset}\label{ssec:dataset}\bench{} uses Cholec80 (80 videos, 7 phases and 7 tools) and CholecT50 (50 videos with phase labels and 100 triplet classes), sampled at 1\,fps. After resolving shared recordings, the training, validation and test splits contain 53/11/21 procedures spanning 32.5/6.1/15.5\,h, respectively. The benchmark contains 11{,}667 interaction instances (7{,}315/1{,}320/3{,}032 across the splits), including 1{,}893 FP/PA instances with annotation-derived response windows for evaluating answer content and timing. \par Phase-transition alerts use $[b_k,b_k+10\,\mathrm{s}]$, and preparation reminders use $[b_k-60\,\mathrm{s},b_k]$. Duration, first-appearance and count answers use tolerances of 20\%, 10\,s and $\pm1$, respectively. Progress questions use S-F1. CM matches phase boundaries and triplet events within 30\,s and requires narration within 10\,s after each phase.\subsection{Implementation Details}
\label{ssec:impl}
We fine-tune OneStreamer-4B~\cite{zeng2026onestreamer}, built on Qwen3-VL-4B~\cite{bai2025qwen3vl}, using LoRA (rank 16) on NVIDIA RTX 6000 Ada GPUs. Training comprises 1 epoch of memory writing and 2 epochs of joint interaction training. We use AdamW with an initial learning rate of $10^{-4}$, cosine decay and a global batch size of 32. We set $W=32$, $\kappa=3$\,s, $\epsilon=5$\,s and $\eta=10$\,s, and use greedy decoding. Each FP/PA instance and \tok{Summary} write point receives two additional training windows. We drop memory records, jitter timestamps and replace elements.\par Fine-tuned OneStreamer matches our training data, window sampling and epochs. Fine-tuned Qwen3-VL uses the same training split and LoRA setting with 5\,s periodic querying. Offline VLMs use causal frame sampling, periodic querying and model-specific inputs. All methods share task definitions, temporal matching and scoring rules. We use Qwen3-VL-4B-Instruct~\cite{bai2025qwen3vl}, Qwen3.5-4B~\cite{qwen2026qwen35}, Gemma~4-E4B-it~\cite{gemma2026gemma4}, MedGemma~1.5-4B-it~\cite{sellergren2026medgemma15}, Hulu-Med-4B~\cite{jiang2025hulumed}, EndoChat-13B~\cite{wang2025endochat} and SurgVidLM-7B~\cite{wang2025surgvidlm}. For MedVLM, we use the uAI-NEXUS-MedVLM-1.0a-7B-RL checkpoint officially released with \mbox{MedGRPO}~\cite{su2026medgrpo}.

\subsection{Comparison with State-of-the-Art Methods}
Table~\ref{tab:main} reports zero-shot and fine-tuned comparisons. Among zero-shot VLMs, MedVLM achieves the best RT accuracy (55.35\%) and Avg$_\mathrm{M}$ (17.59\%), yet all zero-shot models score below 3\% on FP and PA and below 7\% on CM.
\par \textit{Periodic evaluation.} Under 5\,s polling, \ours{} exceeds fine-tuned Qwen3-VL by 2.92/3.26 points in Avg$_\mathrm{M}$/Avg$_\mathrm{S}$. It leads on 10 of 12 metrics, raising FP from 21.84\% to 27.43\%, while fine-tuned Qwen3-VL remains higher on PA and narration (S5).
\par \textit{Proactive evaluation.} With autonomous timing, \ours{} achieves Avg$_\mathrm{M}$/Avg$_\mathrm{S}$ of 42.64\%/48.57\%, exceeding fine-tuned OneStreamer by 6.00/5.87 points and improving all 12 metrics. The largest mode gains are on FP (+12.26) and CM (+10.06). Documentation (S4) gains most (+12.96). Compared with its periodic variant, proactive \ours{} achieves higher BM, CM and Avg$_\mathrm{S}$ without external prompts. Fig.~\ref{fig:qualitative} contrasts our accurate recall and timely replies with OneStreamer's premature phase-start response. Overall, \ours{} achieves the best aggregate scores under both periodic and proactive evaluation, benefiting from workflow anchoring.

\subsection{Efficiency Analysis}
In a separate profiling experiment, we compare \ours{} with OneStreamer on two full-length test procedures at 1\,fps (Fig.~\ref{fig:eff}). \ours{} reduces peak context, median time to first token (TTFT) of question calls and peak GPU memory by 46.9\%, 28.6\% and 10.2\%, respectively. WAM summarizes completed-phase observations, reducing historical text while retaining the recent visual window. Mean update time is 0.53\,s excluding frame I/O, below the 1\,s input interval. These results show that WAM keeps the context compact over long procedures, enabling \ours{} to process full-length surgical videos in real time.

\subsection{Ablation Study}
We ablate WAM and DTL against fine-tuned OneStreamer (Table~\ref{tab:main}), which uses free-caption memory and sparse state supervision (PSTL)~\cite{zeng2026onestreamer}, with the same data and training schedule. As shown in Table~\ref{tab:ablation}, introducing WAM improves BM by 3.36 and CM by 7.68 percentage points, DTL improves FP by 9.36 and CM by 7.18 points, and combining both performs best overall. WAM brings the larger gain on BM and PA, whereas DTL brings the larger gain on FP. Adding DTL to WAM further improves FP by 8.60 points, while adding WAM to DTL raises BM by 2.16 points. The combination performs best on all modes except PA, where WAM alone is higher (17.58\% \emph{vs.} 16.64\%). Together, WAM strengthens retrospective recall and DTL enables timely responses, confirming that the two modules are complementary for streaming surgical interaction.

\section{Conclusion}
\label{sec:conclusion}
We introduced \ours{} for streaming surgical video understanding. WAM consolidates timestamped observations into phase summaries, providing compact, queryable surgical memory. DTL resamples target response times within valid clinical windows to guide timely proactive interaction. We further built \bench{} to evaluate streaming surgical interaction. Experiments show gains in memory and forward proactive QA with lower context and inference costs. Evaluation is currently limited to laparoscopic cholecystectomy. Future work will improve response timing and evaluate generalization to other surgical procedures.

\clearpage 


{\small
\let\originalthebibliography\thebibliography
\renewcommand{\thebibliography}[1]{%
  \originalthebibliography{#1}%
  \setlength{\itemsep}{0pt}%
  \setlength{\parsep}{0pt}%
  \setlength{\parskip}{0pt}%
  \renewcommand{\newblock}{\unskip\space\ignorespaces}%
  \microtypesetup{expansion=true}%
  \tolerance=2000
  \parfillskip=0pt plus .75\linewidth
}
\bibliographystyle{IEEEbib}
\bibliography{refs}

@article{zeng2026onestreamer,
  author  = {Xiangyu Zeng and Yuandong Yang and Zhiqiu Zhang and Yuhan Zhu and Xinhao Li and Qingyi Si and Dingyu Yao and Changlian Ma and Haoran Chen and Xinyu Chen and others},
  title   = {{OneStreamer}: Unifying perception, memory, and proactive response in streaming video interaction},
  journal = {arXiv preprint arXiv:2610.01762},
  year    = {2026}
}

@inproceedings{he2026surgonair,
  author    = {Jingyi He and Yue Zhou and Long Bai and Kun Yuan and Nassir Navab and Yuan Bi},
  title     = {{SurgOnAir}: Hierarchy-aware real-time surgical video commentary},
  booktitle = {MICCAI},
  pages     = {573--582},
  year      = {2026}
}

@article{twinanda2017endonet,
  author  = {Andru P. Twinanda and Sherif Shehata and Didier Mutter and Jacques Marescaux and Michel de Mathelin and Nicolas Padoy},
  title   = {{EndoNet}: A deep architecture for recognition tasks on laparoscopic videos},
  journal = {IEEE Trans. Med. Imaging},
  volume  = {36},
  number  = {1},
  pages   = {86--97},
  year    = {2017}
}

@article{nwoye2022rendezvous,
  author  = {Chinedu Innocent Nwoye and Tong Yu and Cristians Gonzalez and Barbara Seeliger and Pietro Mascagni and Didier Mutter and Jacques Marescaux and Nicolas Padoy},
  title   = {Rendezvous: Attention mechanisms for the recognition of surgical action triplets in endoscopic videos},
  journal = {Med. Image Anal.},
  volume  = {78},
  pages   = {102433},
  year    = {2022}
}

@inproceedings{seenivasan2022surgicalvqa,
  author    = {Lalithkumar Seenivasan and Mobarakol Islam and Adithya K. Krishna and Hongliang Ren},
  title     = {{Surgical-VQA}: Visual question answering in surgical scenes using transformer},
  booktitle = {MICCAI},
  pages     = {33--43},
  year      = {2022}
}

@article{yuan2024ssgvqa,
  author  = {Kun Yuan and Manasi Kattel and Jo{\"e}l L. Lavanchy and Nassir Navab and Vinkle Srivastav and Nicolas Padoy},
  title   = {Advancing surgical {VQA} with scene graph knowledge},
  journal = {Int. J. Comput. Assist. Radiol. Surg.},
  volume  = {19},
  number  = {7},
  pages   = {1409--1417},
  year    = {2024}
}

@inproceedings{he2024pitvqa,
  author    = {Runlong He and Mengya Xu and Adrito Das and Danyal Z. Khan and Sophia Bano and Hani J. Marcus and Danail Stoyanov and Matthew J. Clarkson and Mobarakol Islam},
  title     = {{PitVQA}: Image-grounded text embedding {LLM} for visual question answering in pituitary surgery},
  booktitle = {MICCAI},
  pages     = {488--498},
  year      = {2024}
}

@article{li2024llavasurg,
  author  = {Jiajie Li and Garrett Skinner and Gene Yang and Brian R. Quaranto and Steven D. Schwaitzberg and Peter C. W. Kim and Jinjun Xiong},
  title   = {{LLaVA-Surg}: Towards multimodal surgical assistant via structured surgical video learning},
  journal = {arXiv preprint arXiv:2408.07981},
  year    = {2024}
}

@article{bai2025qwen3vl,
  author  = {Shuai Bai and Yuxuan Cai and Ruizhe Chen and Keqin Chen and Xionghui Chen and Zesen Cheng and Lianghao Deng and Wei Ding and Chang Gao and Chunjiang Ge and others},
  title   = {{Qwen3-VL} technical report},
  journal = {arXiv preprint arXiv:2511.21631},
  year    = {2025}
}

@article{wang2025internvl35,
  author  = {Weiyun Wang and Zhangwei Gao and Lixin Gu and Hengjun Pu and Long Cui and Xingguang Wei and Zhaoyang Liu and Linglin Jing and Shenglong Ye and Jie Shao and others},
  title   = {{InternVL3.5}: Advancing open-source multimodal models in versatility, reasoning, and efficiency},
  journal = {arXiv preprint arXiv:2508.18265},
  year    = {2025}
}

@article{jiang2025hulumed,
  author  = {Songtao Jiang and Yuan Wang and Sibo Song and Tianxiang Hu and Chenyi Zhou and Bin Pu and Yan Zhang and Zhibo Yang and Yang Feng and Joey Tianyi Zhou and others},
  title   = {{Hulu-Med}: A transparent generalist model towards holistic medical vision-language understanding},
  journal = {arXiv preprint arXiv:2510.08668},
  year    = {2025}
}

@inproceedings{chen2024videollmonline,
  author    = {Joya Chen and Zhaoyang Lv and Shiwei Wu and Kevin Qinghong Lin and Chenan Song and Difei Gao and Jia-Wei Liu and Ziteng Gao and Dongxing Mao and Mike Zheng Shou},
  title     = {{VideoLLM-online}: Online video large language model for streaming video},
  booktitle = {CVPR},
  pages     = {18407--18418},
  year      = {2024}
}

@inproceedings{wang2024mmduet,
  author    = {Yueqian Wang and Xiaojun Meng and Yuxuan Wang and Jianxin Liang and Jiansheng Wei and Huishuai Zhang and Dongyan Zhao},
  title     = {{VideoLLM} knows when to speak: Enhancing time-sensitive video comprehension with video-text duet interaction format},
  booktitle = {Findings of EMNLP},
  pages     = {6338--6359},
  year      = {2025}
}

@inproceedings{qian2025dispider,
  author    = {Rui Qian and Shuangrui Ding and Xiaoyi Dong and Pan Zhang and Yuhang Zang and Yuhang Cao and Dahua Lin and Jiaqi Wang},
  title     = {Dispider: Enabling video {LLMs} with active real-time interaction via disentangled perception, decision, and reaction},
  booktitle = {CVPR},
  pages     = {24045--24055},
  year      = {2025}
}

@inproceedings{chen2025livecc,
  author    = {Joya Chen and Ziyun Zeng and Yiqi Lin and Wei Li and Zejun Ma and Mike Zheng Shou},
  title     = {{LiveCC}: Learning video {LLM} with streaming speech transcription at scale},
  booktitle = {CVPR},
  pages     = {\mbox{29083--29095}},
  year      = {2025}
}

@inproceedings{xu2025streamingvlm,
  author    = {Ruyi Xu and Guangxuan Xiao and Yukang Chen and Liuning He and Kelly Peng and Yao Lu and Song Han},
  title     = {{StreamingVLM}: Real-time understanding for infinite video streams},
  booktitle = {ICLR},
  pages     = {61463--61475},
  year      = {2026}
}

@article{sellergren2026medgemma15,
  author  = {Andrew Sellergren and Chufan Gao and Fereshteh Mahvar and Timo Kohlberger and Fayaz Jamil and Madeleine Traverse and Alberto Tono and Bashir Sadjad and Lin Yang and Charles Lau and others},
  title   = {{MedGemma 1.5} technical report},
  journal = {arXiv preprint arXiv:2604.05081},
  year    = {2026}
}

@article{gemma2026gemma4,
  author  = {{Gemma Team}},
  title   = {{Gemma 4} technical report},
  journal = {arXiv preprint arXiv:2607.02770},
  year    = {2026}
}

@misc{qwen2026qwen35,
  author = {{Qwen Team}},
  title  = {{Qwen3.5}: Towards Native Multimodal Agents},
  month  = feb,
  year   = {2026},
  howpublished = {\url{https://qwen.ai/blog?id=qwen3.5}}
}

@article{wang2025endochat,
  author  = {Guankun Wang and Long Bai and Junyi Wang and Kun Yuan and Zhen Li and Tianxu Jiang and Xiting He and Jinlin Wu and Zhen Chen and Zhen Lei and others},
  title   = {{EndoChat}: Grounded multimodal large language model for endoscopic surgery},
  journal = {Med. Image Anal.},
  volume  = {107},
  pages   = {103789},
  year    = {2026}
}

@inproceedings{su2026medgrpo,
  title     = {{MedGRPO}: Multi-task reinforcement learning for heterogeneous medical video understanding},
  author    = {Su, Yuhao and Choudhuri, Anwesa and Gao, Zhongpai and Planche, Benjamin and Nguyen, Van Nguyen and Zheng, Meng and Shen, Yuhan and Innanje, Arun and Chen, Terrence and Elhamifar, Ehsan and Wu, Ziyan},
  booktitle = {CVPR},
  pages     = {2788--2798},
  year      = {2026}
}

@inproceedings{wang2025surgvidlm,
  author    = {Guankun Wang and Junyi Wang and Wenjin Mo and Long Bai and Kun Yuan and Ming Hu and Jinlin Wu and Junjun He and Yiming Huang and Nicolas Padoy and others},
  title     = {{SurgVidLM}: Towards multi-grained video understanding with large language model in robot-assisted surgery},
  booktitle = {ICRA},
  pages     = {13177--13184},
  year      = {2026},
  doi       = {10.1109/ICRA57385.2026.11697585}
}

@article{maierhein2017sds,
  author  = {Lena Maier-Hein and Swaroop S. Vedula and Stefanie Speidel and Nassir Navab and Ron Kikinis and Adrian Park and Matthias Eisenmann and Hubertus Feussner and Germain Forestier and Stamatia Giannarou and others},
  title   = {Surgical data science for next-generation interventions},
  journal = {Nat. Biomed. Eng.},
  volume  = {1},
  number  = {9},
  pages   = {691--696},
  year    = {2017}
}

@article{maierhein2022sds,
  author  = {Lena Maier-Hein and Matthias Eisenmann and Duygu Sarikaya and Keno M{\"a}rz and Toby Collins and Anand Malpani and Johannes Fallert and Hubertus Feussner and Stamatia Giannarou and Pietro Mascagni and others},
  title   = {Surgical data science -- from concepts toward clinical translation},
  journal = {Med. Image Anal.},
  volume  = {76},
  pages   = {102306},
  year    = {2022}
}
}
\end{document}